\pdfoutput=1
\documentclass[11pt]{article}

\usepackage[final]{acl}

\usepackage{times}
\usepackage{latexsym}
\usepackage{listings}   % 导入代码显示相关包
\usepackage{tcolorbox}  % 导入tcolorbox包
\tcbuselibrary{listings, skins}  % 启用tcolorbox的listings库和外观调整库
\usepackage[T1]{fontenc}
\usepackage[utf8]{inputenc}

\usepackage{microtype}

\usepackage{inconsolata}
\usepackage{algorithmicx}
\usepackage{microtype}
\usepackage{booktabs} % for professional tables
\usepackage{hyperref}

\usepackage{amssymb}
\usepackage{amsthm}
\usepackage{amsmath}
\usepackage{amssymb}
\usepackage{mathtools}

\newcommand{\E}{\mathop{\mathbb{E}}\displaylimits}

\usepackage{makecell}
\usepackage{bbm}
\usepackage[textsize=tiny]{todonotes}
\usepackage[capitalize,noabbrev]{cleveref}
\usepackage{inconsolata}
\usepackage{pythonhighlight}
\usepackage[linesnumbered,ruled,vlined]{algorithm2e}
\usepackage{xcolor}
\usepackage{fontawesome}
\usepackage{multirow}
\usepackage{booktabs}
\usepackage{adjustbox}
\usepackage{enumitem}
\usepackage{graphicx}
\usepackage{subfigure}
\usepackage{xcolor}
\usepackage{colortbl}
\usepackage{xcolor}
\usepackage{pgf}
\usepackage{pgfplotstable} % For handling and displaying tables
\usepackage{pgfplots} % For calculations and color mappings
\usepackage{colortbl} % For coloring table cells
\usepackage{booktabs} % For better-looking tables
\usepackage{etoolbox}
\usepackage{makecell}
\usepackage{courier}
\usepackage{authblk}
\usepackage{twemojis}
\newcommand{\sysname}{\textbf{T}$^\star$}

\title{\sysname: Progressive Block Scaling for Masked Diffusion Language Models \\ Through Trajectory Aware Reinforcement Learning}

\author{
{\bf Hanchen Xia$^{\dagger^\star}$,
Baoyou Chen$^{^\dagger\diamondsuit\star}$,
Yutang Ge$^\ddagger$,
Guojiang Zhao$^\mathsection$,} \\ 
{\bf Siyu Zhu$^{\dagger\spadesuit\diamondsuit}$\thanks{Corresponding author.}} \\
$^{\dagger}$Shanghai Academy of AI for Science, \\
$^{\spadesuit}$ Shanghai Innovation Institute, $^{\diamondsuit}$Fudan University, \\
$^{\ddagger}$School of Mathematical Sciences, Shanghai Jiao Tong University, \\
$^{\mathsection}$Carnegie Mellon University, \\
\texttt{\{xiahanchen, chenbaoyou\}@sais.org.cn}
}

\begin{document}
\maketitle

\begingroup
\renewcommand{\thefootnote}{\(\star\)}
\footnotetext{Equally contributed to this work}
\endgroup
% \def\thefootnote{*}\footnotetext{Equally contributed to this work.}\def\thefootnote{\arabic{footnote}}

% \begin{abstract}
% We reveal how trajectory-wise reward assignment in reinforcement learning shapes the denoising schedule of masked diffusion language models (MDMs).
% We further propose a simple progressive training strategy \sysname. The plot below \siyu{In the following figure caption or the main text, give a definition of the ``Base'', ''TraeRL`` with reference, ``\sysname''. Some description of the following two chart.} shows the proposed strategy scales the block size of \texttt{SDAR-1.7B-Chat} with minimal performance degradation.
% This curriculum mitigates the degradation that typically arises when increasing block size \siyu{the degradation only observed from the AR to discrete diffusion training scheme? or both for training from scratch?}, allowing the model to retain strong reasoning ability inherited from autoregressive LLMs while benefiting from diffusion-style decoding with higher parallelism and a more flexible schedule \siyu{token prediction schedule}.

% % \par\smallskip
% % \begin{center}
% %   \includegraphics[width=\linewidth]{tstar_small.pdf}
% % \end{center}
% \end{abstract}

\begin{abstract}
We present \sysname, a simple \textsc{TraceRL}-based curriculum for progressive block-size scaling in masked diffusion language models (MDMs).
Starting from an AR-initialized small-block MDM, \sysname\ gradually increases the block size while re-optimizing the denoising policy at each stage, enabling higher-parallelism decoding with limited degradation on math reasoning benchmarks.
Across two SDAR scales and three benchmarks, \sysname\ consistently outperforms direct large-block \textsc{TraceRL} and is substantially more stable during training.
Our schedule analysis suggests that the learned policy does not simply revert to a strictly left-to-right order; instead, it retains block-size-specific non-monotone updates while improving accuracy.
\end{abstract}

\section{Introduction}

Before the current wave of large language models (LLMs), bidirectional Transformers trained with masked language modeling were a widely adopted backbone for NLP systems, with BERT and its optimized variants as canonical examples \cite{devlin-etal-2019-bert,liu2019roberta}.
Today, autoregressive (AR) modeling via next-token prediction dominates both scaling practice and deployed systems \cite{brown2020gpt3,touvron2023llama}.
% A key reason is its strong causal inductive bias: natural language is conventionally produced as an ordered sequence, and modeling the joint distribution through a left-to-right factorization reduces generation to repeated next-token prediction \cite{bengio2003nnlm}.

% However, human reasoning is often not well captured by a single linear chain: it branches, revisits intermediate hypotheses, and jumps between partial states.
% Recent prompting and inference-time frameworks explicitly operationalize such non-linear structures, for example by exploring trees and graphs over intermediate ``thought'' units rather than committing to a single chain \cite{yao2023tot,DBLP:conf/aaai/BestaBKGPGGLNNH24,yao-etal-2024-got}.

In parallel, diffusion language models have begun to emerge as viable alternatives or complements to the autoregressive decoding paradigm.
Masked diffusion models stochastically mask a subset of tokens under a ratio-parameterized corruption process and optimize cross-entropy on masked positions to recover the original sequence \cite{sahoo2024mdlm}.
For scalability, recent work initializes diffusion LMs from pretrained autoregressive LLMs and trains them with random-mask diffusion objectives \cite{ye2025dream7b,cheng2025sdar}.
At inference time, they often adopt blockwise decoding that denoises tokens within each block while generating blocks autoregressively to preserve global coherence \cite{arriola2025blockdiffusion}.
In this setting, the block size is a control parameter that interpolates between stronger AR-like causality and higher-parallel masked updates.

Within each block, the denoising schedule is typically determined by model confidence.
Given a prompt $Q$ and the current partially denoised sequence $x^{(s)}$, let $\mathcal{M}^{(s)}$ denote the set of masked positions at denoising step $s$.
For each $i\in\mathcal{M}^{(s)}$, the model predicts a token distribution over the vocabulary $\mathcal{V}$.
A common heuristic defines the confidence score
\begin{equation}
\begin{aligned}
& c_i^{(s)} = \max_{v\in\mathcal{V}} p_\theta(x_i=v \mid x^{(s)}, Q),
\quad i\in\mathcal{M}^{(s)}, \\
& U^{(s)} =\{ i\in\mathcal{M}^{(s)} : c_i^{(s)}\ge\eta\},
\end{aligned}
\label{eq:confidence_schedule}
\end{equation}
where $\eta\in(0,1)$ is a confidence threshold that controls how many tokens are finalized at each step. Then materializes tokens in $U^{(s)}$ (e.g., via argmax or sampling), while leaving the rest masked for subsequent refinement.
% We can view masked diffusion models (MDMs) as relaxing the strong inductive bias of autoregressive language models: instead of committing to a fixed left-to-right factorization, an MDM may choose its own unmasking order, while also enabling higher-parallel generation.
% \siyu{The preceding sentence and the following sentence are not logically related.}
% However, a larger block size further reduces the decoding constraint, which can pose additional challenges for logical reasoning. Empirically, 

When examining the \textbf{\texttt{SDAR}} series models across scales (1.7B--30B) and block sizes (4--64), we find that math-centric reasoning becomes increasingly sensitive to larger blocks: accuracy generally degrades as block size $B$ grows, with more pronounced drops for smaller models, which is also reported by \citet{cheng2025sdar}.
% \begin{figure}[h]
%   \centering
%   \includegraphics[width=1.05\linewidth]{larger_block.pdf}
%   \caption{Accuracy on MATH500 (left) and \textsc{MathBench} (right) as a function of block size across model scales, using a confidence-threshold schedule with $\tau=0.8$.}
%   \label{fig:larger-block}
% \end{figure}
We consider the standard \emph{absorbing-state} corruption used in masked diffusion LMs: once a token is replaced by the special \texttt{[/MASK]} symbol in the forward corruption process, subsequent corruption steps keep that position as \texttt{[/MASK]} rather than replacing it with another token.
Under this corruption process, maximizing the ELBO yields a denoising objective; equivalently, the negative-ELBO reduces to a reweighted cross-entropy over masked positions:
\begin{equation}
\begin{aligned}
& \mathcal{L}(\theta)
=
\mathbb{E}_{x_0 \sim p_{\rm data}, x_t \sim q(x_t | x_0), t \sim \mathrm{U}(0,1)} \\
& \Bigg[
-\frac{1}{t}\sum_{\ell=1}^{L}
\mathbf{1}_{\{x_{t,\ell}=\text{{\texttt{[/MASK]}}}\}}\,
\cdot\log p_\theta(x_{0,\ell}\mid x_t)
\Bigg],
\label{eq:mdm_random_mask_loss}
\end{aligned}
\end{equation}
Here, \(x_0\) denotes the clean target response and \(x_t\) its corrupted version at masking ratio \(t\).
For a block of size \(B\), the expected number of \texttt{[/MASK]} positions is \(tB\), so larger blocks contain more masked tokens to resolve within each denoising stage.
Since we scale block sizes in powers of two (\(B=2^n\)), the number of jointly unresolved positions per stage increases with the stage index \(n\).
Standard supervised fine-tuning (SFT) data specify only the final token targets \(x_0\)—that is, which response tokens should ultimately be produced for a prompt—but do not specify an intermediate \emph{unmasking schedule}, namely which subset of positions should be finalized at each denoising step.

In this work, we propose \sysname, an easy-to-implement yet effective strategy for progressive block-size scaling that increases block size with minimal performance degradation.
\sysname~offers a practical route for masked diffusion models (MDMs) to preserve the strong reasoning capability inherited from AR-initialized small-block models while moving toward higher-parallel decoding.
Further analysis suggests that \sysname~can induce an alternative decoding schedule, rather than reverting to the canonical left-to-right schedule.
% This raises the possibility that, if there exists an optimal reasoning schedule ${\rm S}^\star$, MDMs may learn it internally through RL.

\section{Methodology}
\subsection{Trajectory-aware RL}
We adopt \textsc{TraceRL} as our trajectory-aware reinforcement learning backbone, and build our method on top of it \cite{wang2025tracerl}.
TraceRL views diffusion decoding as a multi-step denoising trajectory and performs policy optimization on the same trajectory used at inference.
Given a prompt $Q$, a diffusion LM produces a trajectory
$\tau = \tau(1)\cup\cdots\cup\tau(T)$, where $T$ is the number of denoising steps and $\tau(t)$ denotes the set of tokens decoded (unmasked) at step $t$.
For brevity, we denote the trajectory prefix by $\tau_{<t}:=\tau(1{:}t{-}1)$ and suppress the dependence on $Q$ when it is clear.

We treat each newly finalized token as an action.
Concretely, at denoising step $t$, the policy samples token values for a subset of masked positions that are finalized at this step; we denote an action by $o=(i,\hat{x}_i)$, where $i$ is the finalized position and $\hat{x}_i$ is the sampled token.
Accordingly, $\pi_\theta(o\mid\tau_{<t},Q)$ denotes the probability assigned to choosing $\hat{x}_i$ at position $i$ given the current trajectory prefix and prompt. TraceRL applies a PPO-like objective over all decoded tokens along the trajectory:
\begin{equation}
\begin{aligned}
J(\theta)
&=
\E_{\tau \sim \pi_{\theta_{\mathrm{old}}}}
\Bigg[
\sum_{t=1}^{T}\frac{1}{|\tau(t)|}
\sum_{o\in\tau(t)}
C_{\epsilon}\!\left(\rho_t(o),\,A(o)\right)
\Bigg] \\
&\quad
-\beta\,\mathrm{KL}\!\left(\pi_{\theta}\,\|\,\pi_{\theta_{\mathrm{old}}}\right),
\end{aligned}
\label{eq:tracerl_policy}
\end{equation}
where $C_{\epsilon}(r,A)=\min\{rA,\ \mathrm{clip}(r,1-\epsilon,1+\epsilon)A\}$ is the clipped surrogate and
\begin{equation}
\rho_t(o)
=
\frac{\pi_{\theta}\!\left(o \mid \tau_{<t},Q\right)}
     {\pi_{\theta_{\mathrm{old}}}\!\left(o \mid \tau_{<t},Q\right)}.
\label{eq:tracerl_ratio}
\end{equation}
In the simplest verifiable-reward setting, a single sequence-level reward (e.g., correctness of the final answer) is broadcast to the trajectory and used to form the advantages in Eq.~\ref{eq:tracerl_policy}.

To enable finer credit assignment over denoising steps, \textsc{TraceRL} aggregates token-level rewards (and value predictions) into step-level quantities by averaging within each denoising step, and computes step-wise advantages via TD/GAE \cite{schulman2015gae}.
These step advantages are then assigned back to all tokens decoded at the corresponding step, so that learning signals propagate through the entire denoising trajectory rather than only the final output \cite{lightman2023verify}.
% \siyu{Any further discussion of \textsc{TraceRL}? Seems to a pure description of \textsc{TraceRL}.}

\subsection{Progressive Block Scaling}
We propose \sysname, a stage-wise curriculum that alternates between policy adaptation at a fixed block size and block expansion to the next stage.
The partitions below are applied only to the \emph{response} segment; prompt tokens remain fixed context.

Let the response contain \(L\) token positions indexed by \(\{1,\dots,L\}\).
For a block size \(B\), the \emph{standard} partition is
\begin{equation}
\begin{aligned}
\mathcal{P}_{B}^{(0)}= & \{I_m^{(0)}\}_{m=0}^{M-1}, \\
I_m^{(0)}= & \{mB+1,\dots,\min((m+1)B,L)\},
\end{aligned}
\label{eq:std-partition}
\end{equation}
where \(M=\lceil L/B\rceil\).

To reduce sensitivity to fixed block boundaries, we also use a \emph{shifted} partition with offset \(\Delta=B/2\) (all block sizes in our experiments are even):
\begin{equation}
\mathcal{P}_{B}^{(\Delta)}=\{I_m^{(\Delta)}\}_{m=0}^{M'},
\label{eq:shift-partition}
\end{equation}
where
\begin{equation}
\begin{aligned}
I_0^{(\Delta)}= & \{1,\dots,\min(\Delta,L)\}, \\
I_m^{(\Delta)}= & \{(m-1)B+\Delta+1,\dots, \\
& \min(mB+\Delta,L)\}\ \ (m\ge 1),
\end{aligned}
\label{eq:shift-blocks}
\end{equation}
and \(M'=\max\{0,\lceil (L-\Delta)/B\rceil\}\).
Thus, the first shifted block is a prefix of length \(\Delta\), and every subsequent boundary is moved to the right by \(\Delta\) tokens relative to the standard partition.
The shifted variant changes only the block partition used by blockwise diffusion/\textsc{TraceRL}; it does not modify the underlying tokens or rewards.

At a fixed block size \(B\), one \emph{update step} consists of sampling a rollout batch and applying \textsc{TraceRL} once under the standard partition and once under the shifted partition.
A \emph{stage} consists of \(K_B\) such update steps at the same block size.
Only after finishing the stage do we expand to \(2B\).
We therefore use the terms \emph{update step} and \emph{stage} throughout, rather than \emph{epoch}, to avoid ambiguity.
Algorithm~\ref{alg:tstar_pbs} summarizes the procedure.

\newcommand\mycommfont[1]{\footnotesize\rmfamily\itshape\textcolor{gray}{#1}}
\SetCommentSty{mycommfont}
\SetKwComment{Comment}{// }{}
\SetKwInput{KwIn}{Input}
\SetKwInput{KwOut}{Output}
\newcommand{\Split}{\textsc{Split}}
\newcommand{\TraceRL}{\textsc{TraceRL}}

\begin{algorithm}[t]
    \small
    \SetAlCapSkip{0.5em}
    \setlength\belowcaptionskip{-5pt}
    \caption{\sysname: Progressive Block Scaling (stage-wise)}
    \label{alg:tstar_pbs}

    \KwIn{Base model $\theta_0$, dataset $\mathcal{D}$, initial block size $B_0$, target block size $\hat{B}$, updates per stage $K_B$}
    \KwOut{Optimized model $\theta$}

    \Comment{Initialization}
    $\theta \leftarrow \theta_0$;\quad $B \leftarrow B_0$\;

    \While{$B \le \hat{B}$}{
        \Comment{One stage at fixed block size $B$}
        $\Delta \leftarrow B/2$\;
        Construct $\mathcal{P}_{B}^{(0)}$ and $\mathcal{P}_{B}^{(\Delta)}$\;

        \For{$k \leftarrow 1$ \KwTo $K_B$}{
            Sample a rollout batch $d$ from $\mathcal{D}$\;
            $d_1, d_2 \leftarrow \Split(d)$\;

            \Comment{Update under the standard partition}
            $\theta \leftarrow \TraceRL(\theta,\ d_1,\ B,\ \mathcal{P}_{B}^{(0)})$\;

            \Comment{Update under the shifted partition}
            $\theta \leftarrow \TraceRL(\theta,\ d_2,\ B,\ \mathcal{P}_{B}^{(\Delta)})$\;
        }

        \Comment{Expand to the next stage}
        $B \leftarrow 2B$\;
    }
    \KwRet{$\theta$}
\end{algorithm}

\begin{figure*}[h]
  \centering
  \includegraphics[width=0.90\linewidth]{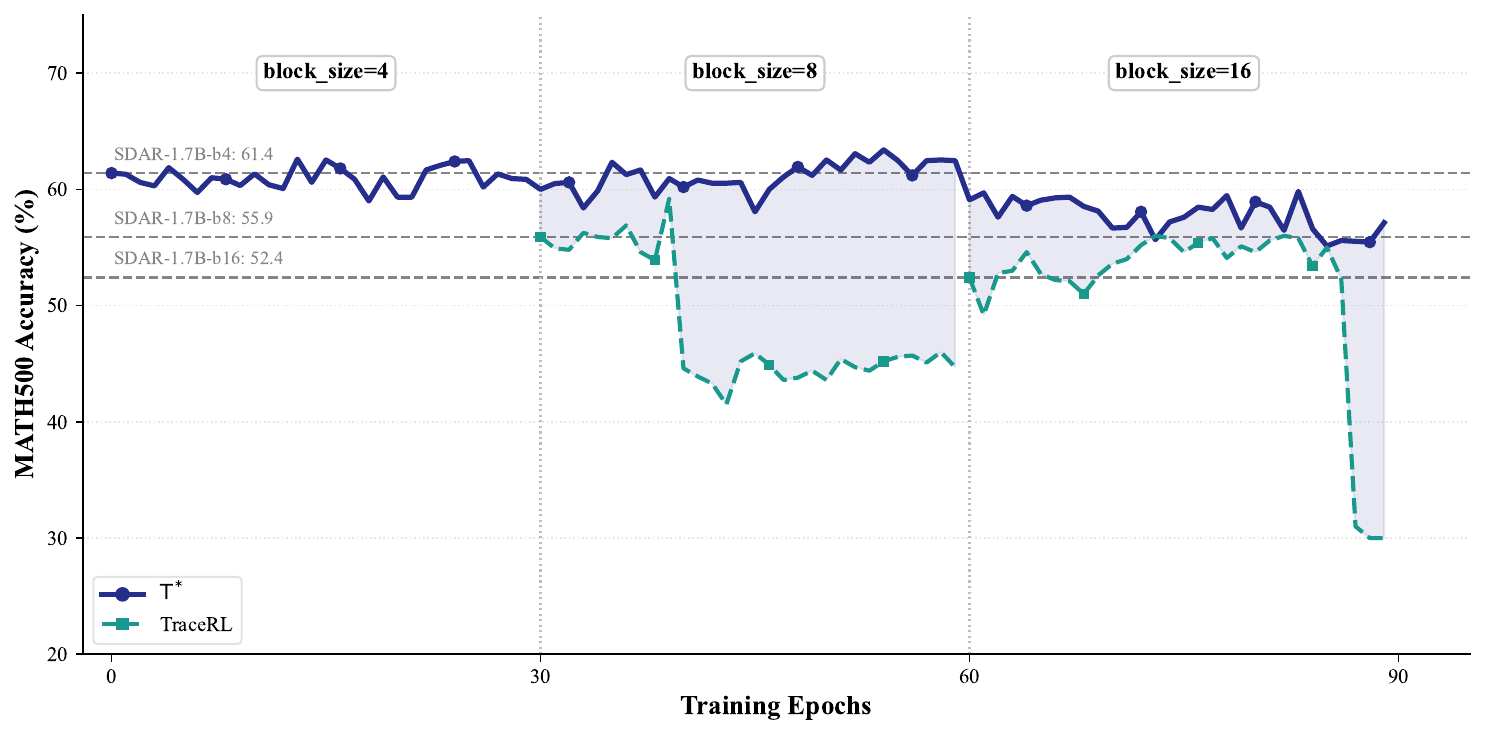}
    \caption{\textbf{Validation accuracy during block scaling (1.7B; x-axis = policy update steps).}
    MATH500 validation accuracy over training update steps for \sysname\ and a direct \textsc{TraceRL} baseline (dashed).
    A \emph{stage} is a contiguous segment of \(K_B\) updates at a fixed block size \(B\).
    Vertical dotted lines indicate stage transitions ($B{=}4 \rightarrow 8 \rightarrow 16$).
    Horizontal dashed lines show the accuracies of the original \textbf{\texttt{SDAR}} checkpoints trained at each block size.}
  \label{fig:tracerl_val_curve}
\end{figure*}

\begin{figure*}[h]
  \centering
  \includegraphics[width=\textwidth]{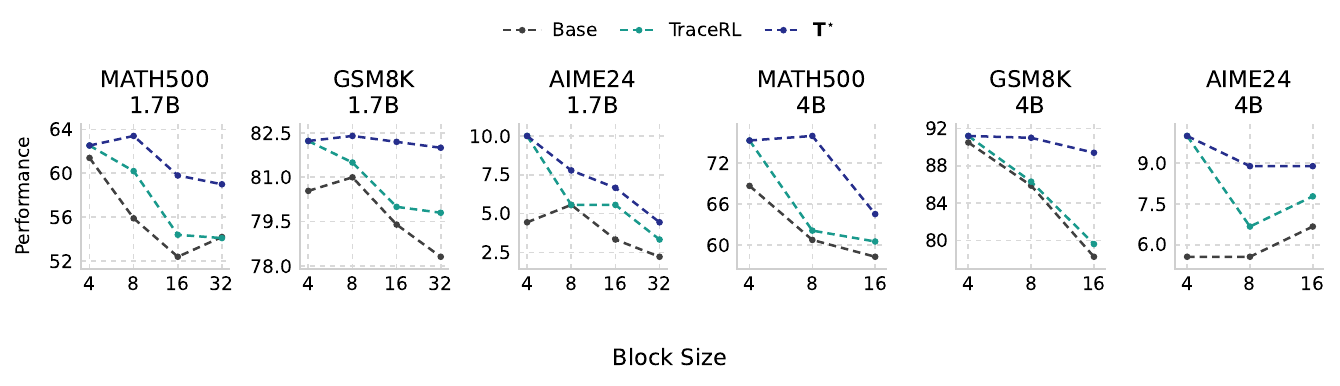}
  \caption{\textbf{Performance vs.\ block size across model scales.}
Performance on MATH500, GSM8K, and AIME24 as a function of block size $B$ for \textbf{\texttt{SDAR}} models at 1.7B (left) and 4B (right).
\textbf{Base} denotes the original \texttt{SDAR-$\cdot$-Chat-b$B$} checkpoint.
\textsc{TraceRL} denotes applying \textsc{TraceRL} directly on the Base checkpoint at the same block size $B$.
\sysname\ denotes our progressive curriculum that warm-starts from a small-block policy and increases $B$ stage-by-stage (Alg.~\ref{alg:tstar_pbs}).}
  \label{fig:tstar-full}
\end{figure*}
\section{Experiments}
\subsection{Setup}
We conduct experiments with the masked diffusion models \textbf{\texttt{SDAR-1.7B-Chat}} and \textbf{\texttt{SDAR-4B-Chat}}. These models are trained via block-diffusion strategy using different block sizes $B \in \{4, 8, 16, 32\}$.
Our training dataset consists of 8K high-quality mathematical problems with difficult levels 3-5 from Openr1math. In each step, we randomly sample 128 problems from the dataset and generate 16 responses per problem using the static sampling strategy. Training process is performed on an 8-GPU H200 cluster using the AdamW optimizer with a learning rate of $1\times 10^{-6}$. To prevent the policy from collapsing or drifting far from the base model, we apply a KL-divergence penalty with $\beta=0.01$. 

\paragraph{Baselines.}
For each target block size \(B\in\{8,16,32\}\) of \textbf{\texttt{SDAR}}, we directly apply 30 \textsc{TraceRL} policy-update steps at that block size.

\paragraph{Evaluation.}
We evaluate on MATH500 \cite{hendrycks2021math,lightman2023lets}, GSM8K \cite{cobbe2021gsm8k}, and AIME24 \cite{aops2024aimeI,aops2024aimeII} and report Pass@3.
Each checkpoint is decoded with its target block size, i.e., the same block size used in the corresponding training stage.

\paragraph{Checkpoint selection.}
For each run, we periodically evaluate intermediate checkpoints on MATH500 and select a single checkpoint per run using MATH500 Pass@3.
We then report that same checkpoint on MATH500, GSM8K, and AIME24.
This protocol follows common practice in RL post-training, where one benchmark is often used for model selection; however, because MATH500 also appears in our headline results, the reported MATH500 scores should be interpreted as model-selection-aware rather than as a strictly held-out estimate.
We therefore focus on cross-benchmark consistency, especially on GSM8K and AIME24, when comparing methods.

\subsection{General Performance}
Figure~\ref{fig:tracerl_val_curve} plots MATH500 validation accuracy throughout training for the 1.7B model.
While \sysname~remains relatively stable across stages, the direct \textsc{TraceRL} baseline exhibits abrupt collapses: a sharp drop occurs during the $B{=}8$ stage (from $\sim$56\% to the low-40\% range), and another collapse appears near the end of the $B{=}16$ stage (down to $\sim$30\%).
We find this instability is highly sensitive to initialization at the target block size: applying \textsc{TraceRL} directly on the \textbf{\texttt{SDAR-1.7B-Chat-b8}} checkpoint collapses, whereas continuing \textsc{TraceRL} at $B{=}8$ starting from a \textsc{TraceRL}-trained $B{=}4$ diffusion policy (our stage transition) remains stable.
A plausible explanation is that larger-block \textbf{\texttt{SDAR}} checkpoints operate under weaker conditioning contexts (cf.\ Eq.~\ref{eq:mdm_random_mask_loss}) and thus start from a lower-confidence regime, yielding noisier rollouts and higher-variance advantage estimates; when such advantages are broadcast to many tokens per denoising step, ratio-based updates can trigger likelihood drift and collapse, consistent with the Lazy Likelihood-Displacement ``death spiral'' analysis for GRPO-style training \cite{deng2025grpo_collapse,gao2025soft}.

Figure~\ref{fig:tstar-full} shows that, across different model sizes, \sysname~consistently matches or exceeds the performance of the base models and \textsc{TraceRL} at the same block size on \textsc{MATH500}, GSM8K, and AIME24.
When we expand the block size, the base models generally show a downward trend, while \sysname~remains more stable and achieves the strongest results at most evaluated block sizes; \textsc{TraceRL} often improves over the base model at smaller blocks but is typically below \sysname~at larger blocks.
All scores below are reported for the single checkpoint selected by the protocol in Sec.~3.1.
The exact scores can be found in Table~\ref{tab:sdtstar_block_scaling} and Appendix~\ref{app:full-results}.

\subsection{Schedule}
% \begin{figure*}[t]
%   \centering
%   \includegraphics[width=0.8\textwidth]{tstar_compare_main_v2.pdf}
%   \caption{Case Study}
%   \label{fig:tstar-case-study-main}
% \end{figure*}
% We sample a set of math problems and analyze their learned decoding schedules.
% Figure~\ref{fig:tstar-case-study} shows a representative example: from the heatmaps of first-unmask step indices, the \textsc{TraceRL}-trained model follows a more monotone (closer-to-in-order) schedule, whereas \sysname induces a more non-monotone schedule.
% Despite being closer to an in-order schedule, \textsc{TraceRL} produces an incorrect final answer in this example, while \sysname yields the correct one.

\begin{figure}[t]
  \centering
  \includegraphics[width=\columnwidth]{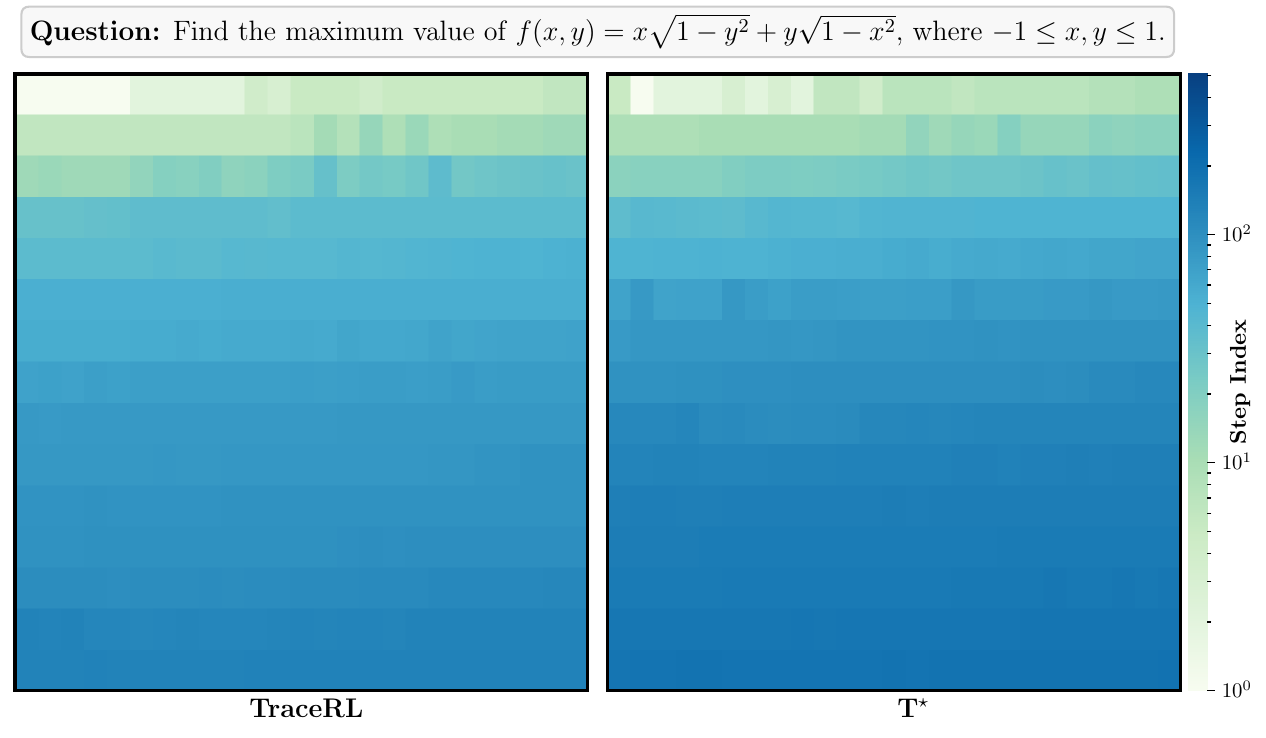}
  \caption{Decoding schedule under \textsc{TraceRL} vs.\ \sysname. More results can be found in Appendix \ref{app:case-study}}
  \label{fig:tstar-heatmap}
\end{figure}

We compute \textsc{LocalStrict} \cite{gong2025diffucoder}.
Let $\pi=(\pi_1,\ldots,\pi_n)$ denote the linearized unmasking order obtained by sorting token positions by their first-unmask step (ties broken by smaller positions).
\textsc{LocalStrict} is defined as the fraction of events that decode the leftmost remaining masked position:
\begin{equation}
\textsc{LocalStrict}
=
\frac{1}{n}\sum_{k=1}^{n}\mathbbm{1}\!\left[\pi_k=\min_{j\ge k}\pi_j\right].
\label{eq:local_strict}
\end{equation}
Higher values indicate a schedule closer to the canonical left-to-right order $S_0$, while lower values reflect more non-monotone masked updates.

\begin{table}[h]
\centering
\small
\setlength{\tabcolsep}{4pt}
\renewcommand{\arraystretch}{1.05}
\begin{tabular}{lccc}
\toprule
\textbf{Model} & \textbf{LocalStrict} &  \textbf{Accuracy}  & TPF \\
\midrule
\textbf{\texttt{Qwen3-1.7B}}          & 1.000 & 70.2 & 1.0 \\
\textbf{\texttt{Qwen2.5-1.5B}}        & 1.000 & 55.0 & 1.0 \\
\midrule
\textbf{\texttt{SDAR-1.7B-b32}}       & 0.743 & 54.2 & 3.74 \\
\quad + \textsc{TraceRL}              & 0.704 & 54.1 & 3.67 \\
\quad + \sysname                      & 0.730 & 59.0 & 3.80 \\
\midrule
\textbf{\texttt{SDAR-1.7B-b16}}       & 0.766 & 52.4 & 3.38 \\
\quad + \textsc{TraceRL}              & 0.824 & 54.4 & 3.41 \\
\quad + \sysname                      & 0.804 & 59.8 & 3.38 \\
\midrule
\textbf{\texttt{SDAR-1.7B-b8}}        & 0.915 & 55.9 & 2.91 \\
\quad + \textsc{TraceRL}              & 0.984 & 60.2 & 2.84 \\
\quad + \sysname                      & 0.854 & 63.4 & 2.95 \\
\bottomrule
\end{tabular}
\caption{\textbf{LocalStrict vs.\ accuracy on \textsc{MATH500} with a decoding-efficiency proxy.}
LocalStrict is computed by Eq.~\ref{eq:local_strict}; higher values indicate a decoding order closer to canonical left-to-right.
TPF denotes \emph{tokens per forward}, i.e., the average number of tokens finalized per model forward pass during decoding (higher implies higher within-block parallelism).}
\label{tab:localstrict_math500}
\end{table}
\paragraph{Decoding efficiency proxy (TPF).}
To illustrate the efficiency benefit of block-size scaling, we report \emph{tokens per forward} (TPF), i.e., the average number of output tokens finalized per model forward pass during decoding (higher is better).
Autoregressive baselines have $\mathrm{TPF}\approx 1$ since they generate one token per forward step, whereas blockwise diffusion can finalize multiple tokens within a block in parallel.
As the block size increases from $B{=}8$ to $B{=}16$ and $B{=}32$, the base \texttt{SDAR-1.7B} model shows a clear increase in TPF (2.91 $\rightarrow$ 3.38 $\rightarrow$ 3.74), corresponding to $\sim$16\% and $\sim$29\% higher TPF, respectively.
Equivalently, for a fixed output length, this reduces the required number of forward passes by $\sim$14\% (from $B{=}8$ to $B{=}16$) and $\sim$22\% (from $B{=}8$ to $B{=}32$).
We stress that TPF is a forward-pass-level proxy rather than a direct end-to-end latency measurement, but it is still informative about the parallelism advantage of larger blocks in forward-pass-limited regimes.
Importantly, applying \textsc{TraceRL} or \sysname\ does not negate this trend: the resulting models retain similar TPF at the same block size, indicating that the reasoning gains from RL-based training are compatible with the parallelism benefits of larger blocks.

Figure~\ref{fig:tstar-heatmap} visualizes token-level first-unmask step indices under \textsc{TraceRL} and \sysname.
Table~\ref{tab:localstrict_math500} reports \textsc{LocalStrict} and accuracy under different block sizes.
Overall, both methods retain largely monotone unmasking behavior (i.e., \textsc{LocalStrict} remains high), but neither collapses to a strictly deterministic left-to-right schedule; instead, the learned step-wise schedules differ under the target block size (see Appendix~\ref{app:case-study} for more examples).

\section{Conclusion}
Experiments show that \sysname~stably scales block size with limited performance degradation, providing a practical recipe for transferring reasoning ability from AR-initialized small-block checkpoints to larger-block diffusion decoding.
We further analyze the instability of direct \textsc{TraceRL} at larger block sizes and relate it to cumulative likelihood drift under noisy rollouts.
Finally, our schedule analysis suggests that the learned denoising policy under a target block size is not simply a reversion to the canonical left-to-right order.

Recent work encourages non-linear reasoning via explicit external scaffolds such as tree-/graph-structured search over intermediate thoughts \cite{yao2023tot,DBLP:conf/aaai/BestaBKGPGGLNNH24,yao-etal-2024-got}.
In contrast, our results suggest that RL can also reshape the model's internal token-finalization order without introducing an external search procedure, making internal schedule learning a complementary direction.

\newpage
\section*{Limitations}
The limitations of this work can be summarized as:
\begin{itemize}
    \item \sysname~mitigates but does not fully eliminate degradation under block expansion; we suspect residual drops are partly due to the lack of a high-quality ``cold-start'' stage.
    \item We did not scale to very large blocks (e.g., $B{=}64$ or $128$) in our \sysname~curriculum, because the inference engine becomes unstable at large block sizes.
\end{itemize}

\section*{Acknowledgments}
This work was supported in part by the Shanghai Municipal Commission of Economy and Informatization (No.~2025-GZL-RGZN-BTBX-01011), the Natural Science Foundation of Shanghai (No.~24ZR1407200), and the Shanghai Oriental Talents Project (No.~QNKJ2024060).

% \section*{Ethical Statements}
% In this paper, we propose strategies to improve the SQL generation capabilities of LLMs.
% To the best of our knowledge, we do not expect our system would have negative impacts on society.

\bibliography{custom}

\onecolumn
\newpage

\appendix
\section{Appendix}
\label{app:appendix}

\subsection{Full Results}
\label{app:full-results}
\begin{table*}[h]
\centering
\small
\setlength{\tabcolsep}{5pt}
\renewcommand{\arraystretch}{1.05}
\begin{tabular}{llccc}
\toprule
\textbf{Base model} & \textbf{Method} & \textbf{MATH500} $\uparrow$ & \textbf{GSM8K} $\uparrow$ & \textbf{AIME24} $\uparrow$ \\
\midrule

\multirow{2}{*}{\rotatebox{0}{\textbf{\texttt{SDAR-1.7B-Chat-b4}}}}
  & --       & 61.40 & 80.54 & 4.44  \\
  & \textsc{TraceRL}  & 62.53 & 82.23 & 10.00 \\
\midrule
\multirow{3}{*}{\textbf{\texttt{SDAR-1.7B-Chat-b8}}}
  & --       & 55.90 & 81.00 & 5.56 \\
  & \textsc{TraceRL}  & 60.20 & 81.50 & 5.56 \\
  & \sysname & 63.40 & 82.40 & 7.78 \\
\midrule
\multirow{3}{*}{\textbf{\texttt{SDAR-1.7B-Chat-b16}}}
  & --       & 52.40 & 79.40 & 3.33 \\
  & \textsc{TraceRL}  & 54.40 & 80.00 & 6.66 \\
  & \sysname & 59.80 & 82.20 & 6.66 \\
\midrule
\multirow{3}{*}{\textbf{\texttt{SDAR-1.7B-Chat-b32}}}
  & --       & 54.20 & 78.31 & 2.22 \\
  & \textsc{TraceRL}  & 54.10 & 79.80 & 3.33 \\
  & \sysname & 59.00 & 82.00 & 4.44 \\
\midrule

\multirow{2}{*}{\textbf{\texttt{SDAR-4B-Chat-b4}}}
  & --       & 68.67 & 90.50 & 5.56 \\
  & \textsc{TraceRL}  & 75.33 & 91.20 & 10.00 \\
\midrule
\multirow{3}{*}{\textbf{\texttt{SDAR-4B-Chat-b8}}}
  & --       & 60.73 & 85.87 & 5.56 \\
  & \textsc{TraceRL}  & 62.10 & 86.30 & 6.67 \\
  & \sysname & 76.00 & 91.00 & 8.89 \\
\midrule
\multirow{3}{*}{\textbf{\texttt{SDAR-4B-Chat-b16}}}
  & --       & 58.26 & 78.24 & 6.67 \\
  & \textsc{TraceRL}  & 60.50 & 79.60 & 7.78 \\
  & \sysname & 64.53 & 89.40 & 8.89 \\

\bottomrule
\end{tabular}
\caption{\textbf{Reasoning performance under different block sizes.}
``--'' denotes the original \texttt{SDAR-$\cdot$-Chat-b$B$} checkpoint, \textsc{TraceRL} applies trajectory-aware RL at the same block size $B$, and \sysname~denotes our progressive block-size scaling.}
\label{tab:sdtstar_block_scaling}
\end{table*}

\newpage
\subsection{Case Study}
\label{app:case-study}
\begin{figure*}[h]
  \centering
  \includegraphics[width=\textwidth]{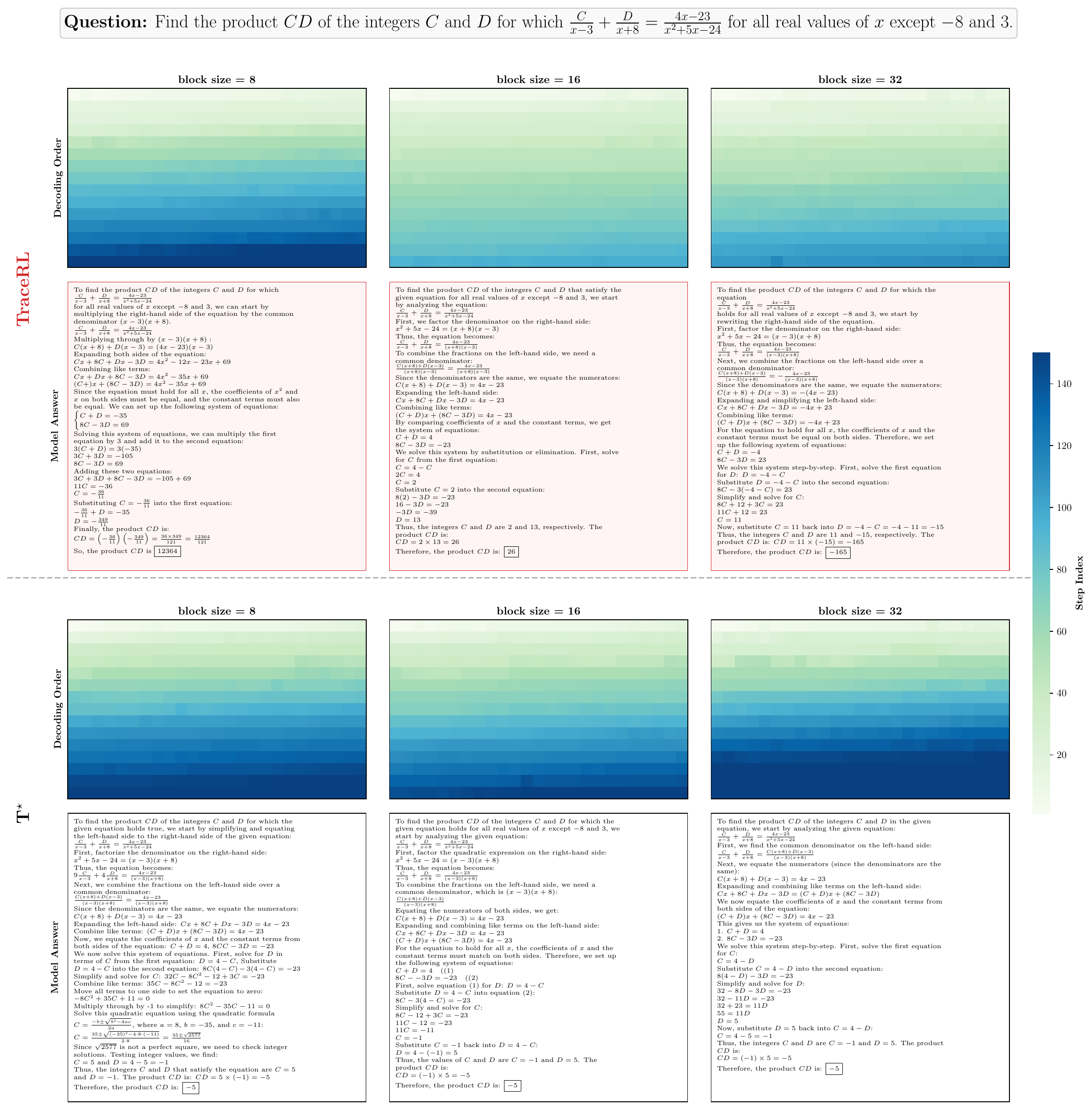}
  \caption{\textbf{Case study: decoding schedule under \textsc{TraceRL} vs.\ \sysname.}
  We visualize the token-level first-unmask step index (heatmaps; darker means decoded later) and the corresponding model solutions for a representative algebra problem, evaluated with block sizes $B\in\{8,16,32\}$.
  The top row shows a model trained with direct \textsc{TraceRL} at the same block size, and the bottom row shows the model obtained by \sysname.}
  \label{fig:tstar-case-study}
\end{figure*}

\end{document}